\pdfoutput=1

\documentclass[11pt]{article}

\usepackage[preprint]{acl}

\usepackage{times}
\usepackage{latexsym}
\usepackage{lingmacros}
\usepackage{tabularray}
\usepackage{graphicx}
\usepackage{subcaption}
\usepackage{rotating} 
\usepackage{dirtytalk}
\usepackage{enumitem}
\usepackage{pdfpages}

\usepackage[T1]{fontenc}

\usepackage[utf8]{inputenc}

\usepackage{microtype}

\usepackage{inconsolata}

\usepackage{graphicx}

\usepackage{tabularx}

\title{Adapting Psycholinguistic Research for LLMs: Gender-inclusive Language in a Coreference Context}

\author{
 \textbf{Marion Bartl\textsuperscript{1,2}}\qquad
 \textbf{Thomas Brendan Murphy\textsuperscript{1,3}}\qquad
 \textbf{Susan Leavy\textsuperscript{1,2}}
\\ 
 \textsuperscript{1}
        Insight SFI Research Centre for Data Analytics \\
 \textsuperscript{2}
        School of Information and Communication Studies \\
 \textsuperscript{3} School of Mathematics and Statistics \\
        University College Dublin\\
\\
 \small{
   \textbf{Correspondence:} \href{mailto:marion.bartl@insight-centre.org}{marion.bartl@insight-centre.org}
 }
}

\begin{document}
\maketitle
\begin{abstract}
Gender-inclusive language is often used with the aim of ensuring that all individuals, regardless of gender, can be associated with certain concepts. While psycholinguistic studies have examined its effects in relation to human cognition, it remains unclear how Large Language Models (LLMs) process gender-inclusive language. Given that commercial LLMs are gaining an increasingly strong foothold in everyday applications, it is crucial to examine whether LLMs in fact interpret gender-inclusive language neutrally, because the language they generate has the potential to influence the language of their users. 
This study examines whether LLM-generated coreferent terms align with a given gender expression or reflect model biases. Adapting psycholinguistic methods from French to English and German, 
we find that in English, LLMs generally maintain the antecedent’s gender but exhibit underlying masculine bias. In German, this bias is much stronger, overriding all tested gender-neutralization strategies.

\end{abstract}

\section{Introduction}

Over the last few decades, activism by feminist linguists has led to increased use of gender-neutral or gender-fair wording, especially in grammatical gender languages such as French or German~\citep{usinger_geschickt_2024, burnett_political_2021}. 
The aim of these forms is to alleviate masculine-default bias and establish representation for people with non-binary gender identities~\citep{freed_women_2020}. Psycholinguistic studies have shown that gender-neutral alternatives can increase the visibility of women and non-binary people~\citep{tibblin_male_2023, fatfouta_unconscious_2023}. 

As Large Language Models (LLMs) are embedded into everyday systems and are used as writing assistants and content creators, the language they generate can have an impact on equal treatment and linguistic representation of women and non-binary people. 
However, while gender bias in NLP is well-researched~\citep{stanczak_survey_2021}, gender-inclusive language in the context of LLMs has only begun to be investigated (\citealp{bartl-leavy-2024-showgirls, watson-etal-2025-language}, a.o.). The processing of gender-inclusive vs. gendered language remains under-explored in English LLMs~\citep{watson_what_2023} and, to our knowledge, entirely unexamined in German LLMs.
To address this, we aim to compare the processing of gendered and gender-inclusive language in both English, a notional gender language, and German, a grammatical gender language.  

We adapt a psycholinguistic study by~\citet{tibblin_male_2023}
to explore how the presence of masculine, feminine or neutral gender in one sentence influences (1) the likelihood of a reference to that gender in a subsequent sentence and (2) the gender mentioned in an LLM-generated completion.
We find that while English LLMs generally keep antecedent and coreferent gender consistent, they are unlikely to use \textit{they} as a singular pronoun and contain underlying masculine bias. The German LLM we tested showed a strong preference for masculine coreferents, regardless of the gender or gender-inclusive strategy used in the antecedent phrase.
We also find evidence that German gender-inclusive language strategies increase the probability of feminine and neutral gender. This finding encourages us to believe that the use of gender-inclusive over generic masculine expressions in German LLMs has the potential to diversify gender representation.

\paragraph{Contributions}
This study translates psycholinguistic methodologies to LLMs, enabling comparisons between human and model reasoning. It introduces a novel approach to assessing whether gender-inclusive expressions promote gender-neutral interpretations within LLMs\footnote{Code and data will be available at \url{https://github.com/marionbartl/GIL-coref-context}}. Additionally, it provides the first analysis of German gender-inclusive strategies in this context, showing that they partially achieve their intended effects by increasing associations with feminine and neutral gender, aligning with psycholinguistic findings.

\section{Background}

The field of \textbf{feminist psycholinguistics} is concerned with evaluating human biases related to language. Studies have shown how masculine generics are in fact not interpreted generically \citep{noll_changes_2018}, and that changing the language to be gender-inclusive also increases mental representation for women and non-binary people~\citep{sato_does_2025, mirabella_role_2024}. 
The term \textit{gender-inclusive language} describes linguistic strategies and neologisms to eliminate male-as-norm bias (\textit{chairman}$\rightarrow$\textit{chairperson}) and emphasize alternative terms that do not reinforce a heteronormative, binary model of gender (\textit{husband/wife}$\rightarrow$\textit{spouse}).

Large Language Models (LLMs) have also been shown to exhibit various social biases, including gender bias~\citep{gupta-etal-2024-sociodemographic}. However, few studies ha explored the \textbf{processing of gender-inclusive language within LLMs}. There are two main areas of investigation: gender-inclusive role nouns (\textit{fire fighter}, \textit{chairperson}, etc.) and gender-neutral pronouns such as singular \textit{they}. The present research addresses both. 

To investigate the processing of \textbf{gender-inclusive role nouns in LLMs}, \citet{watson_what_2023} adapted a psycholinguistic study on sentence acceptability judgments and social attitudes for BERT~\citep{papineau_sally_2022, devlin_bert_2019}. They first calculated BERT's relative probability of a given masculine, feminine or neutral role noun (e.g. \textit{fireman/firewoman/fire fighter}) within a sentence context. BERT's responses were then connected to the social attitudes of the human participants giving the same responses. The researchers found that BERT aligned most with people who had moderate to conservative views.

There are several studies examining \textbf{gender-neutral pronouns in LLMs}. For instance, 
\citet{brandl_how_2022} draw on psycholinguistic research into Swedish neopronouns and adapted an eye-tracking study for LLMs. They demonstrated that while humans do not have trouble processing neopronouns in Swedish~\citep{vergoossen_are_2020}, neopronouns are associated with greater processing difficulty in LLMs. 
Correspondingly, models also have lower pronoun fidelity for feminine and singular \textit{they} pronouns, which means that they are less likely to use them even if they were introduced alongside a corresponding entity \citep{gautam-etal-2024-robust}. 
When comparing an LLM's processing of singular \textit{they} in a generic sense vs. referring to a specific person, models have less trouble with generic \textit{they}~\citep{baumler_recognition_2022}. 
In terms of social attitudes, \citet{watson_what_2023} found that BERT's probabilities for singular \textit{they} resembled the judgments of participants with low to moderate acceptance of non-binary gender.

The psycholinguistic studies that have previously been adapted for LLMs, as well as the research the present study is based on, often contain \textit{anaphora} between two sentences. Anaphora is defined ``in a looser sense, [as] any relation in which something is understood in the light of what precedes it"~\citep{matthews_concise_2014}. The preceding term is called the \textit{antecedent}, while the referring term is the \textit{coreferent}. 
The resolution of this relationship, finding the corresponding antecedent for a coreferent, is a large research field within NLP. \textbf{Coreference Resolution} (CR) is is relevant for downstream NLP tasks such as named entity recognition, summarization or question answering~\citep{liu_brief_2023}. 
CR systems have previously been shown to exhibit gender bias, relying on stereotypes for prediction instead of syntactic information or real-world gender distributions~\citep{rudinger_gender_2018, kotek_gender_2023}. 

To evaluate CR systems for gender biases, challenge datasets based on the Winograd schema~\citep{levesque_winograd_2012} were developed 
~\citep{rudinger_gender_2018,zhao_gender_2018}. These datasets contain instances in which a pronoun must be resolved to refer to one of two previously mentioned entities, such as in the sentence \say{\underline{The paramedic} performed CPR on the passenger even though \textit{she/he/they} knew it was too late.}~\citep{rudinger_gender_2018}. 
While the challenge datasets mostly contain a single sentence, and assess the resolution of pronouns in the singular, this research focuses on coreference between two different sentences in both singular and plural. 

In \textbf{German}, the issue of gender-inclusive language is more intricate than in English. German marks nouns, articles and adjectives for masculine, feminine or neutral gender, traditionally using masculine forms as the generic. A group of individuals of mixed or unknown gender would generally be referred to with masculine gender. 
However, similar to English, masculine generics are interpreted as predominantly masculine~\citep{fatfouta_unconscious_2023}, which is also reflected in NLP models trained on German text~\citep{schmitz_no_2023}.
To increase women's visibility and/or take gender out of the equation, feminist scholars have pushed for linguistic strategies to make role nouns more inclusive~\citep{sczesny_can_2016, dick_gil-galad_2024}. In NLP, there have been efforts to automate the integration of these strategies into text~\citep{amrhein_exploiting_2023}, as well as some research on gender-neutral machine translation into German~\citep{lardelli-etal-2024-gefmt, lardelli-etal-2024-sparks}. However, it is unclear how German gender-fair language is processed by an LLM and we aim to provide some initial answers to this issue in this paper.

\begin{table*}[htbp]
\centering
\small
\begin{tblr}{
  colspec = {X[c,0.1] X[c,0.15] X[l] X[l,0.9]}, 
  row{1} = {c,font=\bfseries}, 
  hlines, 
  rows = {m}
}
lang.  & number &  phrase 1  & phrase 2  \\
\SetCell[r=2]{c} EN  & PL  & The (\textit{sportsmen} | \textit{sportswomen} | \textit{athletes}) were waiting on the steps. & It was obvious that some of the (\textit{men} | \textit{women} | \textit{people}) were in a really good mood.  \\
 & SG  & The (\textit{sportsman} | \textit{sportswoman} | \textit{athlete}) was waiting on the steps. & It was obvious that (\textit{he} | \textit{she} | \textit{they}) (was | were) in a really good mood.  \\
 DE & PL & {Die ( \textit{Tierärzte} |
 \textit{Tierärztinnen} |
 \textit{Tierärztinnen und Tierärzte} |
 \textit{Tierärzte und Tierärztinnen} |
  \textit{TierärztInnen} |
 \textit{Tierärzt*innen} |
 \textit{Tierärzt:innen} |
 \textit{Tierärzt\_innen} ) 
 warteten auf den Stufen.} &	Es war offensichtlich, dass einige (\textit{Männer} | \textit{Frauen} | \textit{Leute}) wirklich guter Laune waren. \\
\end{tblr}
\caption{Examples of antecedent and coreferent combinations for English and German experiments. The templates for English and German are the same, the German antecedents translate to \textit{veterinarian}.}
\label{tab:ante_coref_ex}
\end{table*}

\section{Methodology}
In order to uncover how LLMs process gender-inclusive in contrast to gendered language, we adapted \citeposs{tibblin_male_2023} study design of sentence pairs containing antecedent and coreferent phrases (§\ref{ssec:data}). We used several LLMs (§\ref{ssec:models}) for our experiments on measuring the probability of specific gendered or gender-neutral terms (§\ref{ssec:method_coref_prob}) and analyzing the gender contained in model generations (§\ref{ssec:method_coref_gen}). 

\subsection{Dataset Creation}\label{ssec:data}
We adapted a study design with 44 sentence pairs by~\citet{tibblin_male_2023}. The French sentences in this study designe were translated into English and German using ChatGPT and manually verified.
Each instance in the dataset contains two subsequent phrases. Phrase 1 contains an \textit{antecedent}, a plural noun phrase that is either gendered (\textit{kings}, \textit{au pair girls}) or gender-neutral (\textit{oenologists}, \textit{volunteers}). Phrase 2 contains as the \textit{coreferent} the noun \textit{men} or \textit{women}. 
The content of the phrases can be coherent (\ref{ex:tibblin_data_match})
or incoherent (\ref{ex:tibblin_data_incoh}).

\eenumsentence{
\item \textit{The \textbf{midwives} were entering the hospital. Given the good weather, some of the (\textbf{women}|\textbf{men}) were not wearing jackets.}\label{ex:tibblin_data_match}
\item \textit{The \textbf{referees} were watching the match \underline{in the rain}. Because of \underline{the good weather}, most of the \textbf{men} were wearing shorts.}\label{ex:tibblin_data_incoh}
}\label{ex:tibblin_data}

Using the 11 incoherent instances (cf.~\ref{ex:tibblin_data_incoh}) vs. taking them out had little impact on the outcome of our initial experiments, we therefore retained all 44 instances for experiments measuring coreferent probability. 
Translating the data into English did not always retain the original gender of the antecedent (\textit{Hôtesses de l'air\textsubscript{fem}} -- \textit{flight attendants\textsubscript{neut}}). 
The original data moreover contained imbalanced numbers of gendered/gender-neutral antecedents, which was undesirable for our analysis. We therefore decided to use the data as templates. A template consists of two phrases, the first one with a placeholder for an antecedent, the second with a placeholder for a coreferent. 

\begin{table}[htbp]
\centering
\small
\begin{tblr}{
  colspec = {X[c,0.02]X[l,0.7]X[l]X[l]}, 
  row{1} = {font=\bfseries},
  rows = {m},
  rowsep = {0.8pt}
}
\hline
\# & strategy & DE example & EN translation \\ \hline
1 & masculine & \textit{Akademiker} & academics\textsubscript{masc}\\
2 & feminine & \textit{Akademikerinnen} & academics\textsubscript{fem}\\
3 & coordinated (masc. first) & \textit{Akademiker und Akademikerinnen} & academics\textsubscript{masc} and academics\textsubscript{fem}\\
4 & coordinated (fem. first) & \textit{Akademikerinnen und Akademiker} & academics\textsubscript{fem} and academics\textsubscript{masc}\\
5 & capital I & \textit{AkademikerInnen} & academics\textsubscript{mascFem}\\
6 & colon & \textit{Akademiker:innen} & academics\textsubscript{masc:fem} \\
7 & asterisk & \textit{Akademiker*innen} & academics\textsubscript{masc*fem}\\
8 & underscore & \textit{Akademiker\_innen} & academics\textsubscript{masc\_fem} \\ \hline
\end{tblr}
\caption{Examples of different strategies for gender-inclusive language in German.}
\label{tab:gender_strategies_DE}
\end{table}

\subsubsection{Data for Measuring Coreferent Probability}\label{ssec:data_coref_prob}

\paragraph{English}\label{ssec:data_EN}
Our final English dataset comprises 13,464 instances for the plural (PL) condition and 14,652 instances for the singular (SG) condition. The PL dataset includes 34 antecedent triplets, each paired with three coreferent nouns—\textit{men}, \textit{women}, and \textit{people}—across 44 templates. The SG dataset consists of 37 antecedent triplets, each paired with the pronouns \textit{he}, \textit{she}, and \textit{they}, across 44 templates.
To collect the English antecedents, we utilized gendered terms and their neutral replacements from \citet{bartl-leavy-2024-showgirls}, selecting terms that shared the same neutral equivalent for both masculine and feminine forms (e.g. \textit{swordswoman--swordsman--fencer}). 
Any triplets that were semantically implausible within our template context (e.g., \textit{humankinds}) were manually excluded. 
This resulted in 34 verified triplets for the PL condition and 37 for the SG condition.

\paragraph{German}\label{ssec:data_de}
The final German dataset comprises 10,560 instances, constructed from 10 antecedents, each having eight gender-inclusive variations, paired with three coreferent nouns--\textit{Männer} `\textit{men}', \textit{Frauen} `\textit{women}', and \textit{Personen} `\textit{persons}'--across 44 templates.
To ensure a truly gender-neutral antecedent noun phrase, we maintained coreferent pairs in the plural form, as the German singular inherently marks gender through its article. Instead of translating the English triplets we used professions from the French data to avoid data expansion, given that each antecedent in English had only three variations, whereas German antecedents had eight (Table \ref{tab:DE_ante} in Appendix \ref{app:data}). 
The German gender-inclusive strategies used are outlined in Table \ref{tab:gender_strategies_DE}. They include masculine and feminine forms for reference (strategies 1 and 2), as well as strategies that express both masculine and feminine gender (strategies 3–-5) or incorporate non-binary genders (strategies 6–-8). The latter use characters such as the gender star (*), colon (:), or underscore (\_)~\citep{dick_gil-galad_2024}.

\subsubsection{Data for Coreferent Generation}
In the second set of experiments, we used the models to generate the continuation of Phrase 2 instead of measuring the probability of specific coreferents. The final dataset for coreferent generation comprised 630 instances for English and 160 instances for German. We worked with heavily reduced datasets to minimize annotation workloads and reduce variability in the generations. 
The English dataset was reduced by using the 33 templates with coherent phrases (Example (\ref{ex:tibblin_data_match})) and selecting a reduced set of seven high-frequency plural triplets (Table~\ref{tab:high_freq_EN_ante}). For German, we used the same ten antecedent terms in eight gender variations (§\ref{ssec:data_de}) with 2 coherent templates.

\subsection{Models}\label{ssec:models}

We used six English and one German LLM in the experiments (Table~\ref{tab:model_overview} in Appendix~\ref{app:data}). The models were selected to enable comparison between model sizes and performances.
For the English experiments we used GPT-2~\citep{radford_language_2019} as a baseline, allowing for comparability due to its widespread use in prior research. 
We also tested an adaptation of GPT-2 by~\citet{bartl-leavy-2024-showgirls}, in which the model was fine-tuned with gender-neutral data in order to mitigate gender stereotyping in the model. This model is particularly relevant because our experiments assess how gender-neutral language is processed by LLMs. It can therefore provide insights into how a model that has seen additional gender-neutral language would process gender-neutral language differently. 
We also tested the 1B, 7B and 13B models from the OLMo suite~\citep{groeneveld-etal-2024-olmo}, which are fully open-source, improving transparency for the research community. The different sizes allow us to show the impact of model size on the processing of gendered language.
Qwen2.5 (32B)~\citep{yang_qwen2_2024} was included as our largest model and the best performing pre-trained single-model LLM on the hugg\textbf{}ingface OpenLLM Leaderboard\footnote{\url{https://huggingface.co/spaces/open-llm-leaderboard/open_llm_leaderboard/}} at the time of experimentation (December 2024) within the hardware limitations of our institution.

\subsection{Measuring Coreferent Probability}\label{ssec:method_coref_prob}
We used the LLMs to predict the joint two phrases up to the coreferent (\textit{men}/\textit{women}/\textit{people}), and then obtained the log probability of the coreferent ($\log(p)$) from the probability distribution over the vocabulary. 
For split coreferents, we took the probability of the first component token. Averaging the probabilities of all component tokens would have inflated probabilities, as each component serves as a strong predictor for the subsequent token.

\subsection{Coreferent Generation and Annotation}\label{ssec:method_coref_gen}
We used the models to generate eight tokens for English and ten for German.
The generated continuations were then annotated for gender of the entity mentioned, and whether the mentioned entity was a coreferent of the antecedent in the first sentence.

\paragraph{English} Three annotators were recruited out of a pool of PhD researchers at our institution. Two were native and one was a fluent English speaker. All annotators were paid €60 for 630 items of annotation, each with two labels per item (\texttt{gender} and \texttt{coreference}). The annotation guidelines can be found in Figure \ref{fig:annotation_guidelines} in the Appendix.

Fleiss' kappa was calculated to assess inter-annotator agreement. For the \texttt{gender} labels, the annotations showed $\kappa=0.757$. For the \texttt{coreference} labels, the annotators reached a slightly lower score of $\kappa=0.671$. This is not surprising given that coreference labeling might have been complicated by mentions of several entities or ambiguous phrasing, among others. 
However, both of these scores are in the range of ``substantial agreement'', according to~\citet{landis_measurement_1977}.
We then calculated the final \texttt{gender} and \texttt{coreference} labels based on the majority label. Instances for which all three annotators provided different labels were labeled as \texttt{NULL}. 
There were 22 \texttt{NULL} labels for \texttt{gender} and eight \texttt{NULL} labels for the presence of \texttt{coreference}.

\paragraph{German (pilot)} Due to the lack of German-speaking annotators one of the authors, who is a native speaker of German, annotated the German sentence completions in a pilot experiment. Each completion was annotated for mentioned gender and presence of a coreferent to the antecedent.

\begin{figure*}[ht]
    \centering
    \begin{subfigure}[b]{0.49\textwidth}
        \centering
        \includegraphics[width=\textwidth]{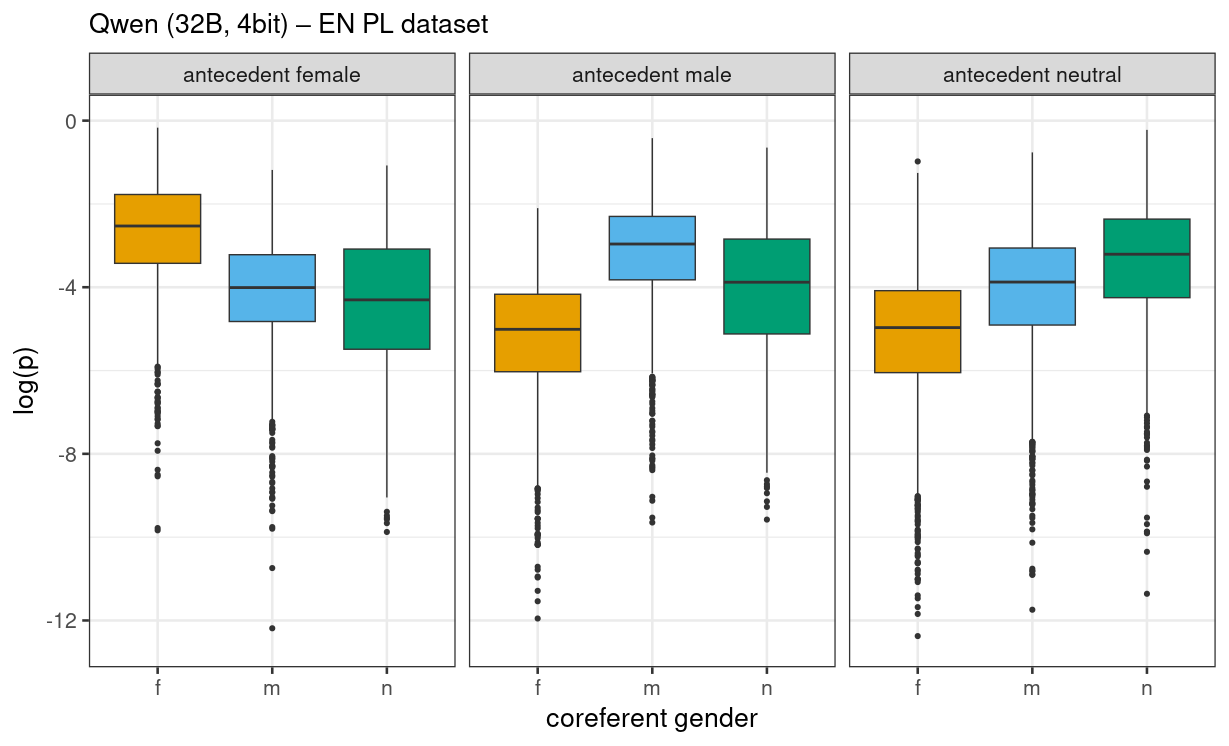}
        \caption{plural}
        \label{fig:qwen_pl}
    \end{subfigure}
    \hfill
    \begin{subfigure}[b]{0.49\textwidth}
        \centering
        \includegraphics[width=\textwidth]{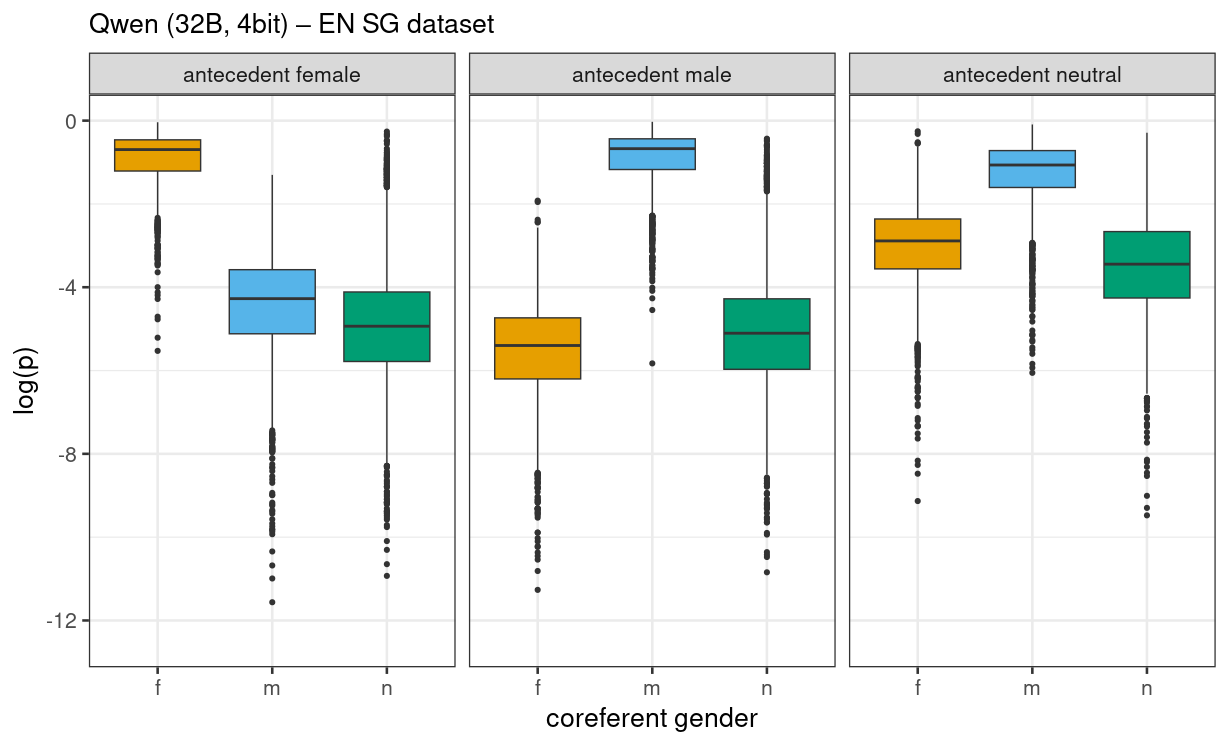}
        \caption{singular}
        \label{fig:qwen_sg}
    \end{subfigure}
    \caption{Distribution of $\log(p)$ of coreferent gender by antecedent gender}
    \label{fig:qwen}
\end{figure*}

\section{Results}

This section lays out the results for our experiments on coreferent probability and coreferent generation. For each of these, we will first present the English and then the German results. 

\subsection{Coreferent Probability}\label{ssec:coref_prob_EN}

\paragraph{English}
For our English results, we provide illustrations for and discuss Qwen-2.5 in detail, as it is the largest and best performing model of those we evaluated. Its results would therefore mirror most closely state-of-the-art models. However, the results for all English models (except the fine-tuned model) follow similar patterns. We provide results and illustrations for the other models, such as the OLMo suite (Figure \ref{fig:olmo}), and the fine-tuned GPT-2 (Figure \ref{fig:gpt_ft}) in Appendix \ref{app:results}.

We performed a two-way ANOVA on the coreferent probabilities produced by Qwen-2.5 (and all other models, cf. Table~\ref{tab:anova_allmodels} in the Appendix), testing the effect of antecedent and coreferent gender on the probability of the coreferent. Effect sizes were labeled following~\citeauthor{field_discovering_2012}'s~\citeyearpar{field_discovering_2012} recommendations.
The ANOVA showed that in the PL setting, the main effect of antecedent gender is statistically significant and small ($F(2, 13455) = 138.59$, $p < .001$; $\eta^2=0.02$, 95\% CI [0.02, 1.00]), which also applied to the main effect of coreferent gender ($F(2, 13455) = 178.33$, $p < .001$; $\eta^2=0.03$, 95\% CI [0.02, 1.00]).
The interaction between antecedent and coreferent gender is statistically significant and large ($F(4, 13455) = 809.94$, $p < .001$; $\eta^2=0.19$, 95\% CI [0.18, 1.00]).
This indicates that in the coreference constructions we are investigating, the probability of the coreferent is most influenced by the correspondence between antecedent and coreferent gender.

Figure \ref{fig:qwen} illustrates the distribution of coreferent probability for the English Qwen-2.5 model in both PL and SG setting. In the PL setting, the model behaves as expected, producing the highest coreferent probability when antecedent gender and coreferent gender correspond (e.g. \textit{The \textbf{bowmen} were going down the street. Some of the \textbf{men} were in a good mood.}). However, for feminine antecedents, masculine coreferents have the second highest probability, indicating masculine bias in the model. The Tukey post-hoc test showed a 21\% lower probability for neutral than masculine coreferents following feminine antecedents (F:N/F:M\footnote{This notation indicates antecedent gender before and coreferent gender after the colon.} = $e^{-0.236}\approx0.79$ , $p<.001$).
This masculine bias is also evident for neutral antecedents. Here, the Tukey post-hoc test showed a probability that was three times higher for masculine than feminine coreferents following neutral antecedents (N:M/N:F = $e^{1.107} \approx 3.03$ , $p<.001$).

The SG setting (Figure \ref{fig:qwen_sg}) is similar to the PL in that matching antecedent and coreferent gender result in the highest probability for masculine and feminine coreferents, for which we used the pronouns \textit{he} and \textit{she}, respectively. Similar to the PL, \textit{he} as a coreferent had a 31\% higher probability than the neutral coreferent \textit{they} for a feminine antecedent (Tukey post-hoc test: F:N/F:M = $e^{-0.37} \approx 0.69$ , $p<.001$), pointing either to masculine bias in the model, or the possibility that singular \textit{they} is not well-recognized or accepted by the LLM. 
This phenomenon can also be observed for neutral antecedents, after which the masculine coreferent \textit{he} has the highest probability, followed by \textit{she} and singular \textit{they}. In fact, the Tukey post-hoc test showed that masculine coreferents following a neutral antecedent had an 88\% higher probability than neutral coreferents (N:N/N:M = $e^{-2.16}\approx0.12$ , $p<.001$).
This result shows that the pronoun \textit{they} is not fully accepted by the model as a singular pronoun. 

\begin{figure*}[ht]
    \centering
    \includegraphics[width=\linewidth]{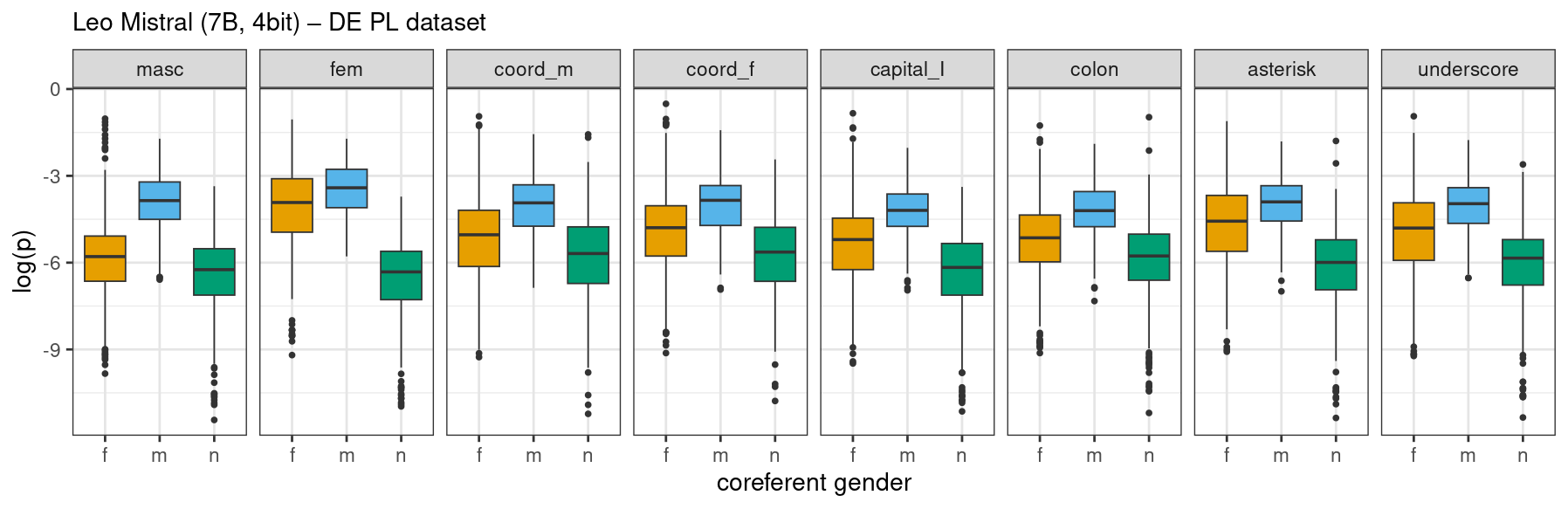}
    \caption{Effect of different gender-inclusive language strategies on coreferent gender probability}
    \label{fig:coref_prob_de}
\end{figure*}

\paragraph{German}
The effects of antecedent gender, coreferent gender, and their interaction on the probability of the coreferent as predicted by Leo Mistral 7B was tested with a two-way ANOVA, as with the English models. Effect sizes were labeled following~\citeauthor{field_discovering_2012}'s~\citeyearpar{field_discovering_2012} recommendations.
The main effect of antecedent gender for the German model is statistically significant and small ($F(7, 10536) =  42.74$, $p<.001$; $\eta^2=0.03$, 95\% CI [0.02, 1.00]), 
and the main effect of coreferent gender is statistically significant and large ($F(2, 10536) = 2601.35$, $p < .001$; $\eta^2=0.33$, 95\% CI [0.32, 1.00]).
The interaction between antecedent and coreferent gender is statistically significant and small ($F(14, 10536) = 36.63$, $p < .001$; $\eta^2=0.05$, 95\% CI [0.04, 1.00]).

In the German ANOVA, contrary to the English results, coreferent gender is the biggest predictor for coreferent probability and not the interaction term. These results become more clear when looking at the probability distributions in Figure \ref{fig:coref_prob_de}:
the masculine continuation \textit{Männer} `men' always shows a much higher probability than \textit{Frauen} `women' and \textit{Personen} `persons'. Therefore, the ANOVA results show coreferent gender to be more predictive than the interaction term.

It can also be seen in Figure~\ref{fig:coref_prob_de} that all German gender-inclusive language strategies lead to an increase in the probability of feminine and gender-neutral coreferents. In the ANOVA results, this finding is supported by the small interaction between antecedent and coreferent gender.
The highest probability for the feminine coreferent can be seen with a feminine antecedent, which is somewhat expected. The second highest probability of a feminine coreferent is brought about by the asterisk strategy, which could be due the feminine PL suffix \textit{-innen} contained in this strategy. However, the capital-I, colon and underscore strategies also contain \textit{-innen}. 
Feminine coreferents generally have the second-highest probability for all gender-inclusive language strategies we tested, meaning that neither strategy favors the generation of \textit{Personen} `persons' as a gender-neutral coreferent. 

\subsection{Coreferent Generation}

\paragraph{English}\label{ssec:coref_gen_EN}
As discussed in Section \ref{ssec:method_coref_gen}, we used majority voting over our three annotation labels to generate the final labels. 
Out of 630 sentence completions, 396 (62.86\%) were labeled as containing a coreferent of the antecedent, 226 (35.87\%) were labeled as not containing a coreferent, and 8 (1.27\%) instances were inconclusive (labeled \texttt{NULL}).

We ran $\chi^2$ tests of independence for both the \texttt{coreference} and \texttt{no-coreference} groups, which were statistically significant ($p<.001$). Effect sizes were labeled following~\citeauthor{funder_evaluating_2019}'s~\citeyearpar{funder_evaluating_2019} recommendations.
In the \texttt{coreference} group, the effect of antecedent gender is very large, ($\chi^2 = 739.57$, $p < .001$; Adjusted Cramer's v = 0.96, 95\% CI [0.90, 1.00]).
In the \texttt{no coreference} group, the effect of antecedent gender is medium ($\chi^2 = 40.12$, $p <.001$; Adjusted Cramer's v = 0.28, 95\% CI [0.16, 1.00]).

\begin{figure}[ht]
    \centering
    \includegraphics[width=\linewidth]{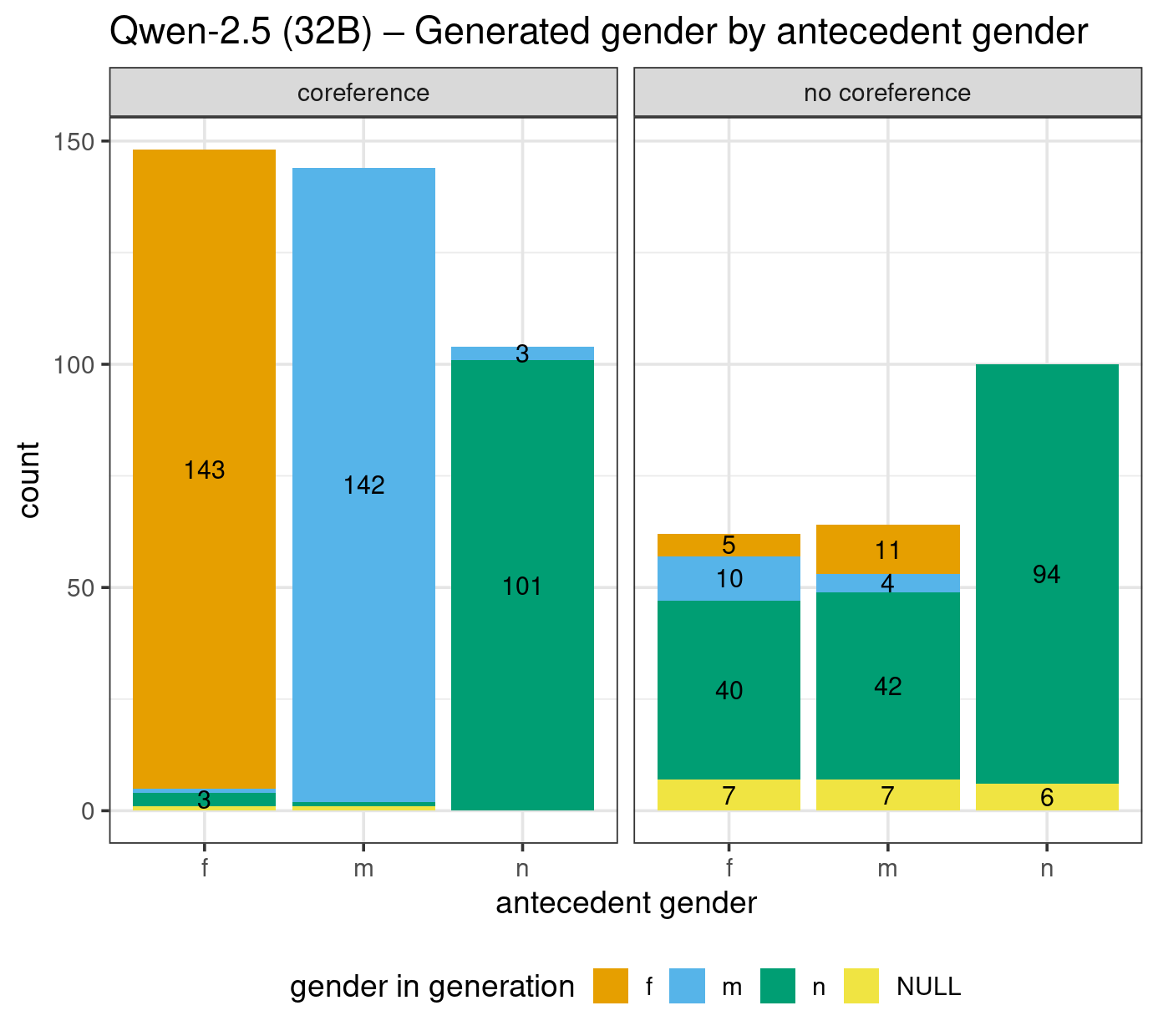}
    \caption{Gender mentioned in the sentence continuation, split by whether or not the generation contains a coreferent of the antecedent}
    \label{fig:coref_pred}
\end{figure}

The distribution of coreferent genders based on antecedent gender and divided by whether or not the continuation contains coreference is illustrated in Figure \ref{fig:coref_pred}.
Figure \ref{fig:coref_pred} shows that if the model generates a coreferent, the coreferent gender follows the antecedent gender with an overwhelming majority.
However, the model generates a coreferent less often when the antecedent is neutral than when it is masculine or feminine. 
In cases where the continuation does not contain a coreferent of the antecedent, the entities mentioned most often have neutral gender. There are also some generations of feminine gender when the antecedent is masculine, and vice versa. This is likely due to prevalence of couplets such as \textit{husband/wife}, meaning that when Phrase 1 mentions \textit{husbands}, Phrase 2 is likely to mention \textit{wives}.

\paragraph{German (pilot)}\label{ssec:coref_gen_DE}

The results for the pilot experiments on German coreferent generation are illustrated in Figure \ref{fig:coref_gen_DE} in Appendix \ref{app:results}. The data are divided into instances where a coreferent noun was generated vs. when there was not. Out of the 160 instances labeled, 100 (62.5\%) contained a coreferent, and 60 (37.5\%) did not. These proportions of generations with and without the coreferent mirror those obtained for English (§\ref{ssec:coref_gen_EN}). 

The Pearson's $\chi^2$ test of independence between antecedent gender and generated coreferent gender suggests that the effect is statistically significant, and very large for the group in which a coreferent was generated ($\chi^2 = 171.79$, $p <.001$; Adjusted Cramer's v = 0.72, 95\% CI [0.56, 1.00]).
For the group in which no coreferent was generated, the $\chi^2$ test also showed a statistically significant and very large effect ($\chi^2 = 70.88$, $p <.001$; Adjusted Cramer's v = 0.54, 95\% CI [0.20, 1.00]).

Figure \ref{fig:coref_gen_DE} shows that similar to the English results (Figure \ref{fig:coref_pred}), masculine and feminine coreferents are mostly generated when the antecedent is masculine or feminine. However, feminine antecedents seem to be a clearer predictor for feminine coreferents, while there are some instances in which a neutral coreferent is generated following a masculine antecedent. 
Generally, Figure \ref{fig:coref_gen_DE} also shows that gender-inclusive antecedents invoke gender-neutral coreferents, which is the intention of using these strategies. One specific case is that coordinated masculine and feminine forms (Table \ref{tab:gender_strategies_DE}, \#3 \& \#4) of the antecedents invoke coordinated coreferents, indicating that the model has a tendency to keep using the same gender form in Phrase 2 that it has seen in Phrase 1.

\section{Discussion}\label{sec:discussion}

Both experiments on measuring coreferent probability and generation of coreferents demonstrated that generally, models tend to match coreferent gender to the antecedent gender. However, there are several caveats to this observation. For English models, whether or not the gender of the coreferent aligns with the antecedent depends on whether the sentences are singular or plural. Our English coreferent probability experiments in the singular setting (Figure \ref{fig:qwen_sg}) showed that when the antecedent is neutral, the masculine pronoun \textit{he} has the highest probability instead of \textit{they}, meaning that models struggle to interpret the pronoun \textit{they} as a singular pronoun. This finding was also reported by~\citet{gautam-etal-2024-robust}.
In language generation applications, this might contribute to the erasure of people of non-binary gender who use \textit{they/them} pronouns, as well as reinforce male-as-norm biases when people of unknown gender are referenced with masculine pronouns~\citep{cao_toward_2021}.

Furthermore, in the English plural experiments the most probable coreferent gender generally follows the gender of the antecedent. However, the second- and third-highest gender probabilities paint a more nuanced picture (Figure \ref{fig:qwen}). For both feminine and neutral antecedents, masculine coreferents are second-most likely. This illustrates bias, because an equitable model would display similar probabilities for feminine and masculine coreferents given a gender-neutral antecedent. For feminine antecedents, it would also assign higher probabilities to neutral over masculine coreferents. Thus, while the model prioritizes gendered context clues--a desirable behavior--it still exhibits an underlying masculine default bias.

This masculine bias was not just underlying but clearly visible in our German experiments. Measuring the probability of specific coreferents showed that \textit{Männer} `\textit{men}' always had a higher probability than either the feminine coreferent \textit{Frauen} `\textit{women}' or neutral coreferent \textit{Personen} `\textit{persons}'. 
This important finding shows that gender bias in the model outweighs information it received in the prompt, which might lead to a reinforcement of male-as-norm bias through a likely prevalence of masculine terms in the output. 
It is important to note, however, that the coreferent generation experiments for German did not show masculine bias to the same extent as the coreferent probability experiments. This might have been due to the model often simply repeating the antecedent phrase in the generations. In our coreferent probability experiments the coreferent terms were different from the antecedent phrases. 

One encouraging finding from the German experiments is that, despite masculine gender having the highest probability, gender-inclusive strategies help increase the probability of feminine and neutral coreferents. This supports one of the aims of using gender-inclusive language: to allow equal association of all genders with respective terms. Our findings clearly illustrate that the model we used does not show this equal association, however, it is promising that the use of gender-fair language can increase the probability of an association with gender-neutral and feminine terms. This finding mirrors the result of psycholinguistic studies into the effects of gender-inclusive language on humans~\citep{tibblin_male_2023, sczesny_can_2016}.

\section{Conclusion}

This research adapted \citeposs{tibblin_male_2023}'s psycholinguistic experiments on the effects on gender-fair language on anaphora resolution to the domain of LLMs. 
We investigated how the use of gendered or gender-inclusive language within one sentence influences the generation of language in consecutive sentences. Our findings indicate that while English LLMs are likely to continue to use the gender of a mentioned entity in a subsequent sentence, there is an underlying prevalence for masculine gender. For German, this bias appears more pronounced, with masculine gender always having the highest probability in spite of feminine or neutral gender information in the previous sentence. 
However, with reference to \citeposs{tibblin_male_2023} findings, gender-inclusive language strategies in German also increase the probability of feminine and gender-neutral referents. This research therefore supports the value of using gender-inclusive language in an LLM context, especially in under-represented languages like German. 

\section{Limitations}

There are several limitations to our work. 
Firstly, the \textbf{types of models} covered mainly included smaller LLMs (1.5\textendash32 billion parameters) due to hardware restrictions at our institution. In contrast, recently released DeepSeek-V3, contains a total of 671B parameters~\citep{deepseek-ai_deepseek-v3_2024}. Future research is needed to determine whether our findings hold for these larger models.

A second limitation is the \textbf{number of coreferents} tested. While we varied the antecedents, we used the same coreferents (PL: \textit{women} (DE: \textit{Frauen}), \textit{men} (DE: \textit{Männer}), \textit{people} (DE: \textit{Personen}); SG: \textit{she}, \textit{he}, \textit{they}). This was done to follow the original setup by~\citet{tibblin_male_2023}. However, in LLMs it would also have been possible to measure the probability of several coreferent candidates. Still, our coreferent generation experiments partially alleviate this bias because they are based on the tokens with the highest probability. 

Finally, we showed how LLMs handle gender-inclusive expressions from one sentence to another. However, LLMs often handle \textbf{longer contexts and exchanges}. Therefore, future research should be conducted in a setting with a longer context. 

\section*{Acknowledgments}
This publication has emanated from research conducted with the financial support of Science Foundation Ireland under Grant number 12/RC/2289\_P2. For the purpose of Open Access, the authors have applied a CC BY public copyright licence to any Author Accepted Manuscript version arising from this submission. 

\bibliography{anthology, zotero}

\begin{thebibliography}{40}
\providecommand{\natexlab}[1]{#1}

\bibitem[{Amrhein et~al.(2023)Amrhein, Schottmann, Sennrich, and Läubli}]{amrhein_exploiting_2023}
Chantal Amrhein, Florian Schottmann, Rico Sennrich, and Samuel Läubli. 2023.
\newblock \href {https://doi.org/10.18653/v1/2023.acl-long.246} {Exploiting {Biased} {Models} to {De}-bias {Text}: {A} {Gender}-{Fair} {Rewriting} {Model}}.
\newblock In \emph{Proceedings of the 61st {Annual} {Meeting} of the {Association} for {Computational} {Linguistics} ({Volume} 1: {Long} {Papers})}, pages 4486--4506, Toronto, Canada. Association for Computational Linguistics.

\bibitem[{Bartl and Leavy(2024)}]{bartl-leavy-2024-showgirls}
Marion Bartl and Susan Leavy. 2024.
\newblock \href {https://doi.org/10.18653/v1/2024.gebnlp-1.18} {From {\textquoteleft}showgirls' to {\textquoteleft}performers': Fine-tuning with gender-inclusive language for bias reduction in {LLM}s}.
\newblock In \emph{Proceedings of the 5th Workshop on Gender Bias in Natural Language Processing (GeBNLP)}, pages 280--294, Bangkok, Thailand. Association for Computational Linguistics.

\bibitem[{Baumler and Rudinger(2022)}]{baumler_recognition_2022}
Connor Baumler and Rachel Rudinger. 2022.
\newblock \href {https://doi.org/10.18653/v1/2022.naacl-main.250} {Recognition of {They}/{Them} as {Singular} {Personal} {Pronouns} in {Coreference} {Resolution}}.
\newblock In \emph{Proceedings of the 2022 {Conference} of the {North} {American} {Chapter} of the {Association} for {Computational} {Linguistics}: {Human} {Language} {Technologies}}, pages 3426--3432, Seattle, United States. Association for Computational Linguistics.

\bibitem[{Brandl et~al.(2022)Brandl, Cui, and Søgaard}]{brandl_how_2022}
Stephanie Brandl, Ruixiang Cui, and Anders Søgaard. 2022.
\newblock \href {https://doi.org/10.18653/v1/2022.naacl-main.265} {How {Conservative} are {Language} {Models}? {Adapting} to the {Introduction} of {Gender}-{Neutral} {Pronouns}}.
\newblock In \emph{Proceedings of the 2022 {Conference} of the {North} {American} {Chapter} of the {Association} for {Computational} {Linguistics}: {Human} {Language} {Technologies}}, pages 3624--3630, Seattle, United States. Association for Computational Linguistics.

\bibitem[{Burnett and Pozniak(2021)}]{burnett_political_2021}
Heather Burnett and Céline Pozniak. 2021.
\newblock \href {https://doi.org/10.1111/josl.12489} {Political dimensions of gender inclusive writing in {Parisian} universities}.
\newblock \emph{Journal of Sociolinguistics}, 25(5):808--831.
\newblock \_eprint: https://onlinelibrary.wiley.com/doi/pdf/10.1111/josl.12489.

\bibitem[{Cao and Daumé(2021)}]{cao_toward_2021}
Yang~Trista Cao and Hal Daumé, III. 2021.
\newblock \href {https://doi.org/10.1162/coli_a_00413} {Toward {Gender}-{Inclusive} {Coreference} {Resolution}: {An} {Analysis} of {Gender} and {Bias} {Throughout} the {Machine} {Learning} {Lifecycle}*}.
\newblock \emph{Computational Linguistics}, 47(3):615--661.

\bibitem[{DeepSeek-AI et~al.(2024)DeepSeek-AI, Liu, Feng, Xue, Wang, Wu, Lu, Zhao, Deng, Zhang, Ruan, Dai, Guo, Yang, Chen, Ji, Li, Lin, Dai, Luo, Hao, Chen, Li, Zhang, Bao, Xu, Wang, Zhang, Ding, Xin, Gao, Li, Qu, Cai, Liang, Guo, Ni, Li, Wang, Chen, Chen, Yuan, Qiu, Li, Song, Dong, Hu, Gao, Guan, Huang, Yu, Wang, Zhang, Xu, Xia, Zhao, Wang, Zhang, Li, Wang, Zhang, Zhang, Tang, Li, Tian, Huang, Wang, Zhang, Wang, Zhu, Chen, Du, Chen, Jin, Ge, Zhang, Pan, Wang, Xu, Zhang, Chen, Li, Lu, Zhou, Chen, Wu, Ye, Ye, Ma, Wang, Zhou, Yu, Zhou, Pan, Wang, Yun, Pei, Sun, Xiao, Zeng, Zhao, An, Liu, Liang, Gao, Yu, Zhang, Li, Jin, Wang, Bi, Liu, Wang, Shen, Chen, Zhang, Chen, Nie, Sun, Wang, Cheng, Liu, Xie, Liu, Yu, Song, Shan, Zhou, Yang, Li, Su, Lin, Li, Wang, Wei, Zhu, Zhang, Xu, Xu, Huang, Li, Zhao, Sun, Li, Wang, Yu, Zheng, Zhang, Shi, Xiong, He, Tang, Piao, Wang, Tan, Ma, Liu, Guo, Wu, Ou, Zhu, Wang, Gong, Zou, He, Zha, Xiong, Ma, Yan, Luo, You, Liu, Zhou, Wu, Ren, Ren, Sha, Fu, Xu, Huang, Zhang, Xie, Zhang, Hao,
  Gou, Ma, Yan, Shao, Xu, Wu, Zhang, Li, Gu, Zhu, Liu, Li, Xie, Song, Gao, and Pan}]{deepseek-ai_deepseek-v3_2024}
DeepSeek-AI, Aixin Liu, Bei Feng, Bing Xue, Bingxuan Wang, Bochao Wu, Chengda Lu, Chenggang Zhao, Chengqi Deng, Chenyu Zhang, Chong Ruan, Damai Dai, Daya Guo, Dejian Yang, Deli Chen, Dongjie Ji, Erhang Li, Fangyun Lin, Fucong Dai, Fuli Luo, Guangbo Hao, Guanting Chen, Guowei Li, H.~Zhang, Han Bao, Hanwei Xu, Haocheng Wang, Haowei Zhang, Honghui Ding, Huajian Xin, Huazuo Gao, Hui Li, Hui Qu, J.~L. Cai, Jian Liang, Jianzhong Guo, Jiaqi Ni, Jiashi Li, Jiawei Wang, Jin Chen, Jingchang Chen, Jingyang Yuan, Junjie Qiu, Junlong Li, Junxiao Song, Kai Dong, Kai Hu, Kaige Gao, Kang Guan, Kexin Huang, Kuai Yu, Lean Wang, Lecong Zhang, Lei Xu, Leyi Xia, Liang Zhao, Litong Wang, Liyue Zhang, Meng Li, Miaojun Wang, Mingchuan Zhang, Minghua Zhang, Minghui Tang, Mingming Li, Ning Tian, Panpan Huang, Peiyi Wang, Peng Zhang, Qiancheng Wang, Qihao Zhu, Qinyu Chen, Qiushi Du, R.~J. Chen, R.~L. Jin, Ruiqi Ge, Ruisong Zhang, Ruizhe Pan, Runji Wang, Runxin Xu, Ruoyu Zhang, Ruyi Chen, S.~S. Li, Shanghao Lu, Shangyan Zhou, Shanhuang
  Chen, Shaoqing Wu, Shengfeng Ye, Shengfeng Ye, Shirong Ma, Shiyu Wang, Shuang Zhou, Shuiping Yu, Shunfeng Zhou, Shuting Pan, T.~Wang, Tao Yun, Tian Pei, Tianyu Sun, W.~L. Xiao, Wangding Zeng, Wanjia Zhao, Wei An, Wen Liu, Wenfeng Liang, Wenjun Gao, Wenqin Yu, Wentao Zhang, X.~Q. Li, Xiangyue Jin, Xianzu Wang, Xiao Bi, Xiaodong Liu, Xiaohan Wang, Xiaojin Shen, Xiaokang Chen, Xiaokang Zhang, Xiaosha Chen, Xiaotao Nie, Xiaowen Sun, Xiaoxiang Wang, Xin Cheng, Xin Liu, Xin Xie, Xingchao Liu, Xingkai Yu, Xinnan Song, Xinxia Shan, Xinyi Zhou, Xinyu Yang, Xinyuan Li, Xuecheng Su, Xuheng Lin, Y.~K. Li, Y.~Q. Wang, Y.~X. Wei, Y.~X. Zhu, Yang Zhang, Yanhong Xu, Yanhong Xu, Yanping Huang, Yao Li, Yao Zhao, Yaofeng Sun, Yaohui Li, Yaohui Wang, Yi~Yu, Yi~Zheng, Yichao Zhang, Yifan Shi, Yiliang Xiong, Ying He, Ying Tang, Yishi Piao, Yisong Wang, Yixuan Tan, Yiyang Ma, Yiyuan Liu, Yongqiang Guo, Yu~Wu, Yuan Ou, Yuchen Zhu, Yuduan Wang, Yue Gong, Yuheng Zou, Yujia He, Yukun Zha, Yunfan Xiong, Yunxian Ma, Yuting Yan, Yuxiang
  Luo, Yuxiang You, Yuxuan Liu, Yuyang Zhou, Z.~F. Wu, Z.~Z. Ren, Zehui Ren, Zhangli Sha, Zhe Fu, Zhean Xu, Zhen Huang, Zhen Zhang, Zhenda Xie, Zhengyan Zhang, Zhewen Hao, Zhibin Gou, Zhicheng Ma, Zhigang Yan, Zhihong Shao, Zhipeng Xu, Zhiyu Wu, Zhongyu Zhang, Zhuoshu Li, Zihui Gu, Zijia Zhu, Zijun Liu, Zilin Li, Ziwei Xie, Ziyang Song, Ziyi Gao, and Zizheng Pan. 2024.
\newblock \href {https://doi.org/10.48550/arXiv.2412.19437} {{DeepSeek}-{V3} {Technical} {Report}}.
\newblock \emph{arXiv preprint}.
\newblock ArXiv:2412.19437 [cs].

\bibitem[{Devlin et~al.(2019)Devlin, Chang, Kenton, and Toutanova}]{devlin_bert_2019}
Jacob Devlin, Ming-Wei Chang, Lee Kenton, and Kristina Toutanova. 2019.
\newblock {BERT}: {Pre}-training of {Deep} {Bidirectional} {Transformers} for {Language} {Understanding}.
\newblock In \emph{Proceedings of {NAACL}-{HLT}}, pages 4171--4186.

\bibitem[{Dick et~al.(2024)Dick, Drews, Pickard, and Pierz}]{dick_gil-galad_2024}
Anna-Katharina Dick, Matthias Drews, Valentin Pickard, and Victoria Pierz. 2024.
\newblock \href {https://aclanthology.org/2024.lrec-main.684} {{GIL}-{GALaD}: {Gender} {Inclusive} {Language} - {German} {Auto}-{Assembled} {Large} {Database}}.
\newblock In \emph{Proceedings of the 2024 {Joint} {International} {Conference} on {Computational} {Linguistics}, {Language} {Resources} and {Evaluation} ({LREC}-{COLING} 2024)}, pages 7740--7745, Torino, Italia. ELRA and ICCL.

\bibitem[{Fatfouta and Sczesny(2023)}]{fatfouta_unconscious_2023}
Ramzi Fatfouta and Sabine Sczesny. 2023.
\newblock \href {https://doi.org/10.1007/s11199-023-01411-8} {Unconscious {Bias} in {Job} {Titles}: {Implicit} {Associations} {Between} {Four} {Different} {Linguistic} {Forms} with {Women} and {Men}}.
\newblock \emph{Sex Roles}, 89(11):774--785.

\bibitem[{Field et~al.(2012)Field, Miles, and Field}]{field_discovering_2012}
Andy Field, Jeremy Miles, and Zoë Field. 2012.
\newblock \emph{Discovering statistics using {R}}.
\newblock Sage publications.

\bibitem[{Freed(2020)}]{freed_women_2020}
Alice~F. Freed. 2020.
\newblock Women, {Language} and {Public} {Discourse}: {Five} decades of sexism and scrutiny.
\newblock In \emph{Innovations and {Challenges}: {Women}, {Language} and {Sexism}}. Routledge.
\newblock Num Pages: 16.

\bibitem[{Funder and Ozer(2019)}]{funder_evaluating_2019}
David~C. Funder and Daniel~J. Ozer. 2019.
\newblock \href {https://doi.org/10.1177/2515245919847202} {Evaluating {Effect} {Size} in {Psychological} {Research}: {Sense} and {Nonsense}}.
\newblock \emph{Advances in Methods and Practices in Psychological Science}, 2(2):156--168.
\newblock Publisher: SAGE Publications Inc.

\bibitem[{Gautam et~al.(2024)Gautam, Bingert, Zhu, Lauscher, and Klakow}]{gautam-etal-2024-robust}
Vagrant Gautam, Eileen Bingert, Dawei Zhu, Anne Lauscher, and Dietrich Klakow. 2024.
\newblock \href {https://doi.org/10.1162/tacl_a_00719} {Robust pronoun fidelity with {E}nglish {LLM}s: Are they reasoning, repeating, or just biased?}
\newblock \emph{Transactions of the Association for Computational Linguistics}, 12:1755--1779.

\bibitem[{Groeneveld et~al.(2024{\natexlab{a}})Groeneveld, Beltagy, Walsh, Bhagia, Kinney, Tafjord, Jha, Ivison, Magnusson, Wang, Arora, Atkinson, Authur, Chandu, Cohan, Dumas, Elazar, Gu, Hessel, Khot, Merrill, Morrison, Muennighoff, Naik, Nam, Peters, Pyatkin, Ravichander, Schwenk, Shah, Smith, Strubell, Subramani, Wortsman, Dasigi, Lambert, Richardson, Zettlemoyer, Dodge, Lo, Soldaini, Smith, and Hajishirzi}]{groeneveld-etal-2024-olmo}
Dirk Groeneveld, Iz~Beltagy, Evan Walsh, Akshita Bhagia, Rodney Kinney, Oyvind Tafjord, Ananya Jha, Hamish Ivison, Ian Magnusson, Yizhong Wang, Shane Arora, David Atkinson, Russell Authur, Khyathi Chandu, Arman Cohan, Jennifer Dumas, Yanai Elazar, Yuling Gu, Jack Hessel, Tushar Khot, William Merrill, Jacob Morrison, Niklas Muennighoff, Aakanksha Naik, Crystal Nam, Matthew Peters, Valentina Pyatkin, Abhilasha Ravichander, Dustin Schwenk, Saurabh Shah, William Smith, Emma Strubell, Nishant Subramani, Mitchell Wortsman, Pradeep Dasigi, Nathan Lambert, Kyle Richardson, Luke Zettlemoyer, Jesse Dodge, Kyle Lo, Luca Soldaini, Noah Smith, and Hannaneh Hajishirzi. 2024{\natexlab{a}}.
\newblock \href {https://doi.org/10.18653/v1/2024.acl-long.841} {{OLM}o: Accelerating the science of language models}.
\newblock In \emph{Proceedings of the 62nd Annual Meeting of the Association for Computational Linguistics (Volume 1: Long Papers)}, pages 15789--15809, Bangkok, Thailand. Association for Computational Linguistics.

\bibitem[{Groeneveld et~al.(2024{\natexlab{b}})Groeneveld, Beltagy, Walsh, Bhagia, Kinney, Tafjord, Jha, Ivison, Magnusson, Wang, Arora, Atkinson, Authur, Chandu, Cohan, Dumas, Elazar, Gu, Hessel, Khot, Merrill, Morrison, Muennighoff, Naik, Nam, Peters, Pyatkin, Ravichander, Schwenk, Shah, Smith, Strubell, Subramani, Wortsman, Dasigi, Lambert, Richardson, Zettlemoyer, Dodge, Lo, Soldaini, Smith, and Hajishirzi}]{groeneveld_olmo_2024}
Dirk Groeneveld, Iz~Beltagy, Evan Walsh, Akshita Bhagia, Rodney Kinney, Oyvind Tafjord, Ananya Jha, Hamish Ivison, Ian Magnusson, Yizhong Wang, Shane Arora, David Atkinson, Russell Authur, Khyathi Chandu, Arman Cohan, Jennifer Dumas, Yanai Elazar, Yuling Gu, Jack Hessel, Tushar Khot, William Merrill, Jacob Morrison, Niklas Muennighoff, Aakanksha Naik, Crystal Nam, Matthew Peters, Valentina Pyatkin, Abhilasha Ravichander, Dustin Schwenk, Saurabh Shah, William Smith, Emma Strubell, Nishant Subramani, Mitchell Wortsman, Pradeep Dasigi, Nathan Lambert, Kyle Richardson, Luke Zettlemoyer, Jesse Dodge, Kyle Lo, Luca Soldaini, Noah Smith, and Hannaneh Hajishirzi. 2024{\natexlab{b}}.
\newblock \href {https://aclanthology.org/2024.acl-long.841} {{OLMo}: {Accelerating} the {Science} of {Language} {Models}}.
\newblock In \emph{Proceedings of the 62nd {Annual} {Meeting} of the {Association} for {Computational} {Linguistics} ({Volume} 1: {Long} {Papers})}, pages 15789--15809, Bangkok, Thailand. Association for Computational Linguistics.

\bibitem[{Gupta et~al.(2024)Gupta, Narayanan~Venkit, Wilson, and Passonneau}]{gupta-etal-2024-sociodemographic}
Vipul Gupta, Pranav Narayanan~Venkit, Shomir Wilson, and Rebecca Passonneau. 2024.
\newblock \href {https://doi.org/10.18653/v1/2024.gebnlp-1.19} {Sociodemographic bias in language models: A survey and forward path}.
\newblock In \emph{Proceedings of the 5th Workshop on Gender Bias in Natural Language Processing (GeBNLP)}, pages 295--322, Bangkok, Thailand. Association for Computational Linguistics.

\bibitem[{Kotek et~al.(2023)Kotek, Dockum, and Sun}]{kotek_gender_2023}
Hadas Kotek, Rikker Dockum, and David Sun. 2023.
\newblock \href {https://doi.org/10.1145/3582269.3615599} {Gender bias and stereotypes in {Large} {Language} {Models}}.
\newblock In \emph{Proceedings of {The} {ACM} {Collective} {Intelligence} {Conference}}, {CI} '23, pages 12--24, New York, NY, USA. Association for Computing Machinery.

\bibitem[{Landis and Koch(1977)}]{landis_measurement_1977}
J.~Richard Landis and Gary~G. Koch. 1977.
\newblock \href {https://doi.org/10.2307/2529310} {The {Measurement} of {Observer} {Agreement} for {Categorical} {Data}}.
\newblock \emph{Biometrics}, 33(1):159--174.
\newblock Publisher: International Biometric Society.

\bibitem[{Lardelli et~al.(2024{\natexlab{a}})Lardelli, Dill, Attanasio, and Lauscher}]{lardelli-etal-2024-sparks}
Manuel Lardelli, Timm Dill, Giuseppe Attanasio, and Anne Lauscher. 2024{\natexlab{a}}.
\newblock \href {https://aclanthology.org/2024.gitt-1.2/} {Sparks of fairness: Preliminary evidence of commercial machine translation as {E}nglish-to-{G}erman gender-fair dictionaries}.
\newblock In \emph{Proceedings of the 2nd International Workshop on Gender-Inclusive Translation Technologies}, pages 12--21, Sheffield, United Kingdom. European Association for Machine Translation (EAMT).

\bibitem[{Lardelli et~al.(2024{\natexlab{b}})Lardelli, Lauscher, and Attanasio}]{lardelli-etal-2024-gefmt}
Manuel Lardelli, Anne Lauscher, and Giuseppe Attanasio. 2024{\natexlab{b}}.
\newblock \href {https://aclanthology.org/2024.eamt-2.19/} {{G}e{FMT}: Gender-fair language in {G}erman machine translation}.
\newblock In \emph{Proceedings of the 25th Annual Conference of the European Association for Machine Translation (Volume 2)}, pages 37--38, Sheffield, UK. European Association for Machine Translation (EAMT).

\bibitem[{Levesque et~al.(2012)Levesque, Davis, and Morgenstern}]{levesque_winograd_2012}
Hector Levesque, Ernest Davis, and Leora Morgenstern. 2012.
\newblock The winograd schema challenge.
\newblock In \emph{Thirteenth {International} {Conference} on the {Principles} of {Knowledge} {Representation} and {Reasoning}}.

\bibitem[{Liu et~al.(2023)Liu, Mao, Luu, and Cambria}]{liu_brief_2023}
Ruicheng Liu, Rui Mao, Anh~Tuan Luu, and Erik Cambria. 2023.
\newblock \href {https://doi.org/10.1007/s10462-023-10506-3} {A brief survey on recent advances in coreference resolution}.
\newblock \emph{Artificial Intelligence Review}, 56(12):14439--14481.

\bibitem[{Matthews(2014)}]{matthews_concise_2014}
P.~H. Matthews. 2014.
\newblock \emph{The concise {Oxford} dictionary of linguistics}, third;3rd; edition.
\newblock Book, Whole. Oxford University Press, Oxford.

\bibitem[{Mirabella et~al.(2024)Mirabella, Mazzuca, Livio, Giannantonio, Rosati, Lorusso, Lingiardi, Borghi, and Giovanardi}]{mirabella_role_2024}
Marta Mirabella, Claudia Mazzuca, Chiara~De Livio, Bianca~Di Giannantonio, Fau Rosati, Maric~Martin Lorusso, Vittorio Lingiardi, Anna~M. Borghi, and Guido Giovanardi. 2024.
\newblock \href {https://doi.org/10.1037/sgd0000729} {The {Role} of {Language} in {Nonbinary} {Identity} {Construction}: {Gender} {Words} {Matter}}.
\newblock \emph{Psychology of sexual orientation and gender diversity}, (Journal Article).
\newblock Publisher: Educational Publishing Foundation.

\bibitem[{Noll et~al.(2018)Noll, Lowry, and Bryant}]{noll_changes_2018}
Jane Noll, Mark Lowry, and Judith Bryant. 2018.
\newblock \href {https://doi.org/10.1007/s10936-018-9577-4} {Changes {Over} {Time} in the {Comprehension} of {He} and {They} as {Epicene} {Pronouns}}.
\newblock \emph{Journal of Psycholinguistic Research}, 47(5):1057--1068.

\bibitem[{Papineau et~al.(2022)Papineau, Podesva, and Degen}]{papineau_sally_2022}
Brandon Papineau, Rob Podesva, and Judith Degen. 2022.
\newblock \href {https://escholarship.org/uc/item/8k59g29g} {‘{Sally} the {Congressperson}’: {The} {Role} of {Individual} {Ideology} on the {Processing} and {Production} of {English} {Gender}-{Neutral} {Role} {Nouns}}.
\newblock \emph{Proceedings of the Annual Meeting of the Cognitive Science Society}, 44(44).

\bibitem[{Radford et~al.(2019)Radford, Wu, Child, Luan, Amodei, and Sutskever}]{radford_language_2019}
Alec Radford, Jeffrey Wu, Rewon Child, David Luan, Dario Amodei, and Ilya Sutskever. 2019.
\newblock Language {Models} are {Unsupervised} {Multitask} {Learners}.

\bibitem[{Rudinger et~al.(2018)Rudinger, Naradowsky, Leonard, and Van~Durme}]{rudinger_gender_2018}
Rachel Rudinger, Jason Naradowsky, Brian Leonard, and Benjamin Van~Durme. 2018.
\newblock \href {https://doi.org/10.18653/v1/N18-2002} {Gender {Bias} in {Coreference} {Resolution}}.
\newblock In \emph{Proceedings of the 2018 {Conference} of the {North} {American} {Chapter} of the {Association} for {Computational} {Linguistics}: {Human} {Language} {Technologies}, {Volume} 2 ({Short} {Papers})}, pages 8--14, New Orleans, Louisiana. Association for Computational Linguistics.

\bibitem[{Sato et~al.(2025)Sato, Gygax, Gabriel, Oakhill, and Escasain}]{sato_does_2025}
Sayaka Sato, Pascal~Mark Gygax, Ute Gabriel, Jane Oakhill, and Lucie Escasain. 2025.
\newblock \href {https://doi.org/10.1525/collabra.128470} {Does {Inclusive} {Language} {Increase} the {Visibility} of {Women}, or {Does} {It} {Simply} {Decrease} the {Visibility} of {Men}? {A} {Missing} {Piece} of the {Inclusive} {Language} {Jigsaw}}.
\newblock \emph{Collabra: Psychology}, 11(1):128470.

\bibitem[{Schmitz et~al.(2023)Schmitz, Schneider, and Esser}]{schmitz_no_2023}
Dominic Schmitz, Viktoria Schneider, and Janina Esser. 2023.
\newblock \href {https://doi.org/10.5070/G6011192} {No genericity in sight: {An} exploration of the semantics of masculine generics in {German}}.
\newblock \emph{Glossa Psycholinguistics}, 2(1).

\bibitem[{Sczesny et~al.(2016)Sczesny, Formanowicz, and Moser}]{sczesny_can_2016}
Sabine Sczesny, Magda Formanowicz, and Franziska Moser. 2016.
\newblock \href {https://doi.org/10.3389/fpsyg.2016.00025} {Can {Gender}-{Fair} {Language} {Reduce} {Gender} {Stereotyping} and {Discrimination}?}
\newblock \emph{Frontiers in Psychology}, 7(Journal Article):25--25.
\newblock Place: Switzerland Publisher: Frontiers Research Foundation.

\bibitem[{Stanczak and Augenstein(2021)}]{stanczak_survey_2021}
Karolina Stanczak and Isabelle Augenstein. 2021.
\newblock \href {http://arxiv.org/abs/2112.14168} {A {Survey} on {Gender} {Bias} in {Natural} {Language} {Processing}}.
\newblock \emph{arXiv preprint}.
\newblock ArXiv:2112.14168 [cs].

\bibitem[{Tibblin et~al.(2023)Tibblin, Granfeldt, van~de Weijer, and Gygax}]{tibblin_male_2023}
Julia Tibblin, Jonas Granfeldt, Joost van~de Weijer, and Pascal Gygax. 2023.
\newblock \href {https://doi.org/10.5070/G60111267} {The male bias can be attenuated in reading: on the resolution of anaphoric expressions following gender-fair forms in {French}}.
\newblock \emph{Glossa Psycholinguistics}, 2(1).

\bibitem[{Usinger and Müller(2024)}]{usinger_geschickt_2024}
Johanna Usinger and Philipp Müller. 2024.
\newblock \href {https://geschicktgendern.de} {Geschickt gendern - das {Genderwörterbuch}}.

\bibitem[{Vergoossen et~al.(2020)Vergoossen, Pärnamets, Renström, and Gustafsson~Sendén}]{vergoossen_are_2020}
Hellen~P. Vergoossen, Philip Pärnamets, Emma~A. Renström, and Marie Gustafsson~Sendén. 2020.
\newblock \href {https://doi.org/10.3389/fpsyg.2020.574356} {Are {New} {Gender}-{Neutral} {Pronouns} {Difficult} to {Process} in {Reading}? {The} {Case} of {Hen} in {SWEDISH}}.
\newblock \emph{Frontiers in Psychology}, 11.
\newblock Publisher: Frontiers.

\bibitem[{Watson et~al.(2023)Watson, Beekhuizen, and Stevenson}]{watson_what_2023}
Julia Watson, Barend Beekhuizen, and Suzanne Stevenson. 2023.
\newblock \href {https://aclanthology.org/2023.acl-long.375} {What social attitudes about gender does {BERT} encode? {Leveraging} insights from psycholinguistics}.
\newblock In \emph{Proceedings of the 61st {Annual} {Meeting} of the {Association} for {Computational} {Linguistics} ({Volume} 1: {Long} {Papers})}, pages 6790--6809, Toronto, Canada. Association for Computational Linguistics.

\bibitem[{Watson et~al.(2025)Watson, Lee, Beekhuizen, and Stevenson}]{watson-etal-2025-language}
Julia Watson, Sophia~S. Lee, Barend Beekhuizen, and Suzanne Stevenson. 2025.
\newblock \href {https://aclanthology.org/2025.coling-main.80/} {Do language models practice what they preach? examining language ideologies about gendered language reform encoded in {LLM}s}.
\newblock In \emph{Proceedings of the 31st International Conference on Computational Linguistics}, pages 1201--1223, Abu Dhabi, UAE. Association for Computational Linguistics.

\bibitem[{Yang et~al.(2024)Yang, Yang, Hui, Zheng, Yu, Zhou, Li, Li, Liu, Huang, Dong, Wei, Lin, Tang, Wang, Yang, Tu, Zhang, Ma, Yang, Xu, Zhou, Bai, He, Lin, Dang, Lu, Chen, Yang, Li, Xue, Ni, Zhang, Wang, Peng, Men, Gao, Lin, Wang, Bai, Tan, Zhu, Li, Liu, Ge, Deng, Zhou, Ren, Zhang, Wei, Ren, Liu, Fan, Yao, Zhang, Wan, Chu, Liu, Cui, Zhang, Guo, and Fan}]{yang_qwen2_2024}
An~Yang, Baosong Yang, Binyuan Hui, Bo~Zheng, Bowen Yu, Chang Zhou, Chengpeng Li, Chengyuan Li, Dayiheng Liu, Fei Huang, Guanting Dong, Haoran Wei, Huan Lin, Jialong Tang, Jialin Wang, Jian Yang, Jianhong Tu, Jianwei Zhang, Jianxin Ma, Jianxin Yang, Jin Xu, Jingren Zhou, Jinze Bai, Jinzheng He, Junyang Lin, Kai Dang, Keming Lu, Keqin Chen, Kexin Yang, Mei Li, Mingfeng Xue, Na~Ni, Pei Zhang, Peng Wang, Ru~Peng, Rui Men, Ruize Gao, Runji Lin, Shijie Wang, Shuai Bai, Sinan Tan, Tianhang Zhu, Tianhao Li, Tianyu Liu, Wenbin Ge, Xiaodong Deng, Xiaohuan Zhou, Xingzhang Ren, Xinyu Zhang, Xipin Wei, Xuancheng Ren, Xuejing Liu, Yang Fan, Yang Yao, Yichang Zhang, Yu~Wan, Yunfei Chu, Yuqiong Liu, Zeyu Cui, Zhenru Zhang, Zhifang Guo, and Zhihao Fan. 2024.
\newblock \href {https://doi.org/10.48550/arXiv.2407.10671} {Qwen2 {Technical} {Report}}.
\newblock \emph{arXiv preprint}.
\newblock ArXiv:2407.10671 [cs].

\bibitem[{Zhao et~al.(2018)Zhao, Wang, Yatskar, Ordonez, and Chang}]{zhao_gender_2018}
Jieyu Zhao, Tianlu Wang, Mark Yatskar, Vicente Ordonez, and Kai-Wei Chang. 2018.
\newblock \href {https://doi.org/10.18653/v1/N18-2003} {Gender {Bias} in {Coreference} {Resolution}: {Evaluation} and {Debiasing} {Methods}}.
\newblock In \emph{Proceedings of the 2018 {Conference} of the {North} {American} {Chapter} of the {Association} for {Computational} {Linguistics}: {Human} {Language} {Technologies}, {Volume} 2 ({Short} {Papers})}, pages 15--20, New Orleans, Louisiana. Association for Computational Linguistics.

\end{thebibliography}

\appendix

\newpage 

\section{Data}\label{app:data}

\begin{table}[htbp]
\centering
\begin{tblr}{
    colspec = {X[l,0.5]XXX}, 
    row{1} = {font=\bfseries,c}, 
    rows = {m,c},
    rowsep=0pt
}
\hline
\textbf{number} & \textbf{neutral} & \textbf{feminine} & \textbf{masculine}\\
\hline
\SetCell[r=7]{c,m} PL & grandparents & grandmothers & grandfathers\\
& monarchs & queens & kings \\
& siblings & sisters & brothers \\
& parents-in-law & mothers-in-law & fathers-in-law \\
& parents & mothers & fathers \\
& children & daughters & sons \\
& spouses & wives & husbands \\ \hline
\end{tblr}
\caption{High frequency English antecedents}
\label{tab:high_freq_EN_ante}
\end{table}

\begin{table}[ht]
\centering
\small
\begin{tblr}{
    colspec = {X[0.4]XX}, 
    row{1} = {font=\bfseries,c}, 
    rows = {m,c},
    column{1} = {font=\bfseries}
}
\hline
lang. & model name & {\# parameters} \\ \hline
\SetCell[r=4]{m} EN & GPT2 & 1.5B\\
& GPT2 fine-tuned & 1.5B \\
& OLMo & 1B, 7B, 13B \\
& Qwen2.5 & 32B \\ \hline
DE & LeoLM Mistral\footnote{\url{https://huggingface.co/jphme/em_german_leo_mistral}} & 7B \\
\hline
\end{tblr}
\caption{Overview of LLMs used}
\label{tab:model_overview}
\end{table}

\begin{table*}[p]
\centering
\small
\begin{turn}{90}

\begin{tblr}{
    colspec = {X[0.02]X[0.8]XXXXXXX}, 
    rows = {m,l},
    row{1} = {font=\bfseries,c}, 
    width=0.95\textheight
}
\hline
\# & masculine & feminine & {coordinated\\feminine first} & {coordinated\\masculine first} & capital I & asterisk & colon & underscore & EN translation\\ \hline
1 & Eigentümer & Eigentümerinnen & Eigentümerinnen und Eigentümer & Eigentümer und Eigentümerinnen & EigentümerInnen & Eigentümer*innen & Eigentümer:innen & Eigentümer\_innen & owners \\
2 & Allergologen & Allergologinnen & Allergologinnen und Allergologen & Allergologen und Allergologinnen & AllergologInnen & Allergolog*innen & Allergolog:innen & Allergolog\_innen & allergists \\
3 & Choreographen & Choreographinnen & Choreographinnen und Choreographen & Choreographen und Choreographinnen & ChoreographInnen & Choreograph*innen & Choreograph:innen & Choreograph\_innen & choreographers \\
4 & Beamte & Beamtinnen & Beamtinnen und Beamte & Beamte und Beamtinnen & BeamtInnen & Beamt*innen & Beamt:innen & Beamt\_innen & civil servants \\
5 & Radfahrer & Radfahrerinnen & Radfahrerinnen und Radfahrer & Radfahrer und Radfahrerinnen & RadfahrerInnen & Radfahrer*innen & Radfahrer:innen & Radfahrer\_innen & cyclists \\
6 & Akademiker & Akademikerinnen & Akademikerinnen und Akademiker & Akademiker und Akademikerinnen & AkademikerInnen & Akademiker*innen & Akademiker:innen & Akademiker\_innen & academics \\
7 & Önologen & Önologinnen & Önologinnen und Önologen & Önologen und Önologinnen & ÖnologInnen & Önolog*innen & Önolog:innen & Önolog\_innen & oenologists \\
8 & Schiedsrichter & Schiedsrichterinnen & Schiedsrichterinnen und Schiedsrichter & Schiedsrichter und Schiedsrichterinnen & SchiedsrichterInnen & Schiedsrichter*innen & Schiedsrichter:innen & Schiedsrichter\_innen & referees \\
9 & Tierärzte & Tierärztinnen & Tierärztinnen und Tierärzte & Tierärzte und Tierärztinnen & TierärztInnen & Tierärzt*innen & Tierärzt:innen & Tierärzt\_innen & veterinarians \\
10 & Archäologen & Archäologinnen & Archäologinnen und Archäologen & Archäologen und Archäologinnen & ArchäologInnen & Archäolog*innen & Archäolog:innen & Archäolog\_innen & archeologists \\
\hline
\end{tblr}
\end{turn}
\caption{German antecedents}
\label{tab:DE_ante}
\end{table*}

\begin{figure*}[ht]
    \centering
    \fbox{\includegraphics[width=\textwidth]{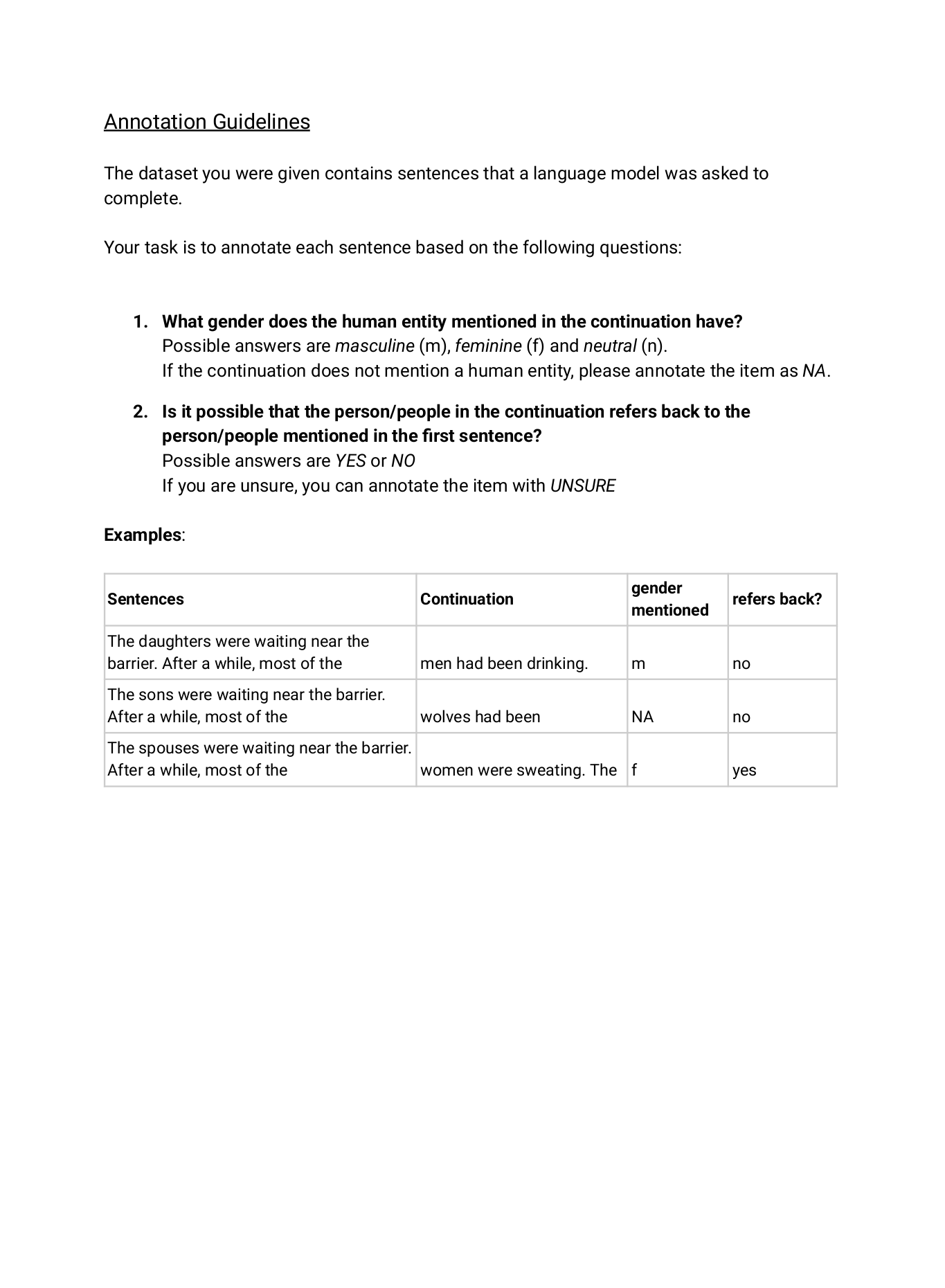}}
    \caption{Annotation guidelines given to annotators for English data}
    \label{fig:annotation_guidelines}
\end{figure*}

\newpage

\section{Results}\label{app:results}

\begin{table*}[ht]
\centering
\small
\begin{tblr}{
    colspec = {X[0.3,l]X[0.3]X[0.3]X[1.2]X[0.3]XXX}, 
    row{1} = {font=\bfseries},
    rows = {m,c},
}
\hline
number & lang. & \# obs. & LLM & quant. & \textit{F}\textsubscript{ante\_gender} & \textit{F}\textsubscript{coref\_gender} & \textit{F}\textsubscript{interaction} \\ \hline
\SetCell[r=7]{m,c} PL & \SetCell[r=6]{m,c} EN & \SetCell[r=6]{m,c} 13464 & GPT-2 & 32bit & 481.6 & 720.2 & 1629.7 \\ 
& & & GPT-2-finetuned & 32bit & 119.8 & 3432.9 & 983.5 \\
& & & OLMo 1B & 4bit & 184.3 & 799.1 & 1011.8 \\
& & & OLMo 7B & 4bit & 67.3 & 142.8 & 720 \\
& & & OLMo 13B & 4bit & 297.8 & 710.4 & 622.8 \\
& & & Qwen 32B & 4bit & 138.6 & 178.3 & 809.9 \\ \hline
& DE & 9240 & EM Leo Mistral 7B & 4bit & 42.74 & 2601.35 & 36.63 \\ \hline \hline
\SetCell[r=6]{m,c} SG & \SetCell[r=6]{m,c} EN & \SetCell[r=6]{m,c} 14652 & GPT-2 & 32bit & 876.6 & 7885.6 & 6336.3 \\
& & & GPT-2-finetuned & 32bit & 111.9 & 44001.9 & 6835.5 \\
& & & OLMo 1B & 4bit & 342.8 & 3998.4 & 4171.4 \\
& & & OLMo 7B & 4bit & 706.3 & 2816.8 & 5509.6 \\
& & & OLMo 13B & 4bit & 592.9 & 3212.2 & 3703.3 \\
& & & Qwen 32B & 4bit & 1231 & 3866 & 4626 \\ \hline
\end{tblr}
\caption{ANOVA effect sizes for antecedent gender, coreferent gender and interaction for all LLMs tested. \\ All effects significant with $p<.001$. \textbf{quant.} = model quantization.}
\label{tab:anova_allmodels}
\end{table*}

\subsection{Model Size Comparison}

\begin{figure*}[ht]
    \centering
    \includegraphics[width=\linewidth]{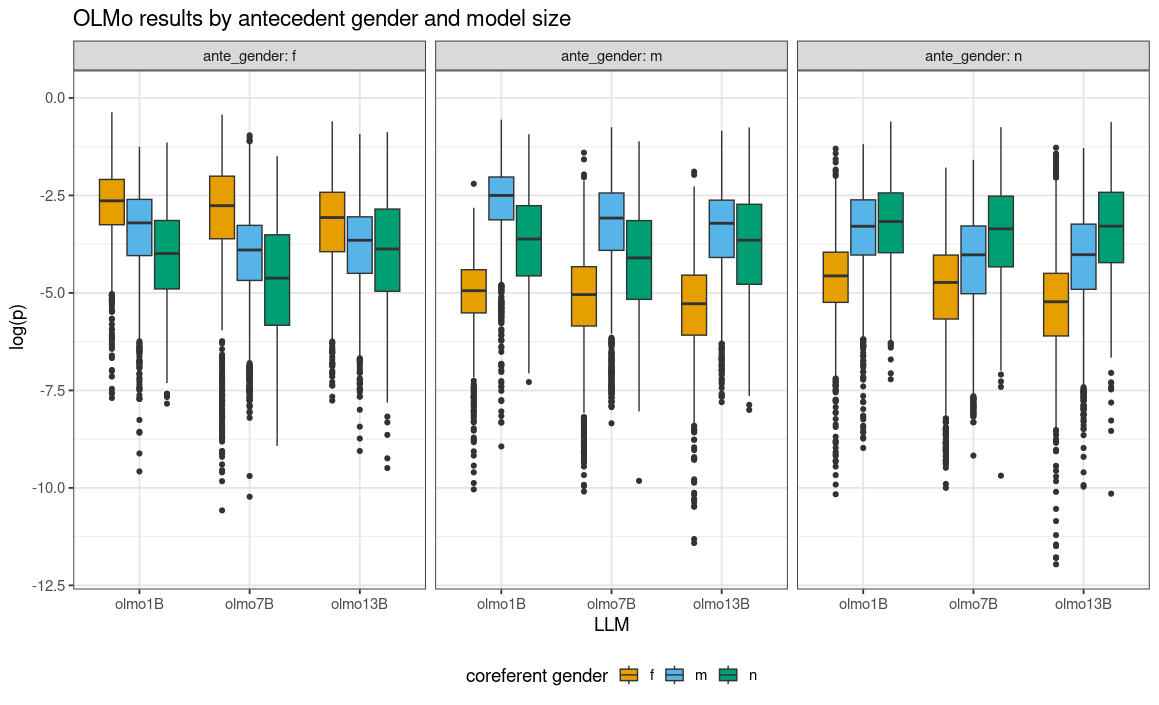}
    \caption{Coreferent probabilities for three OLMo model sizes for feminine, masculine and neutral antecedent gender}
    \label{fig:olmo}
\end{figure*}

Figure \ref{fig:olmo} shows the probability distributions for three OLMo models \citep{groeneveld_olmo_2024} of 1B, 7B and 13B parameters. Overall, the three models show similar distributions for all three antecedent genders that follow those discussed for the Qwen2.5 32B model (Figure \ref{fig:qwen}): the highest probabilities are obtained when antecedent and coreferent gender match, and masculine gender has the second-highest probability for both neutral and feminine antecedent. The probabilities for masculine coreferents across all antecedents are highest for the smallest, 1B parameter model, which could indicate that masculine bias is highest for this model.

\begin{figure*}[ht]
    \centering
    \begin{subfigure}[b]{0.48\textwidth}
        \centering
        \includegraphics[width=\textwidth]{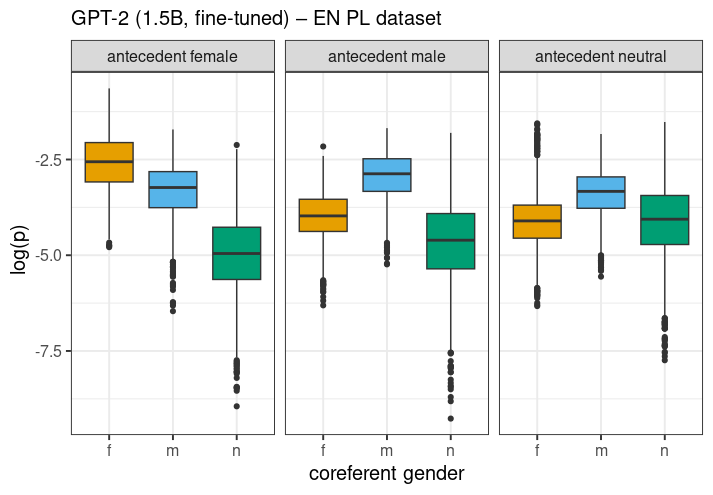}
        \caption{plural}
        \label{fig:gpt_ft_PL}
    \end{subfigure}
    \hfill
    \begin{subfigure}[b]{0.48\textwidth}
        \centering
        \includegraphics[width=\textwidth]{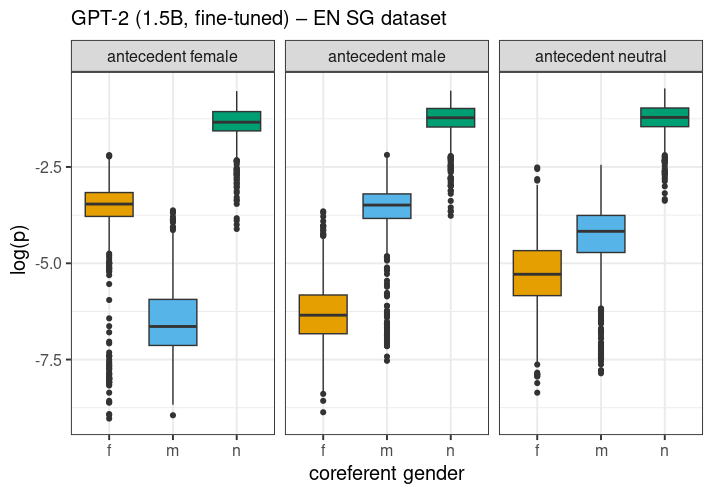}
        \caption{singular}
        \label{fig:gpt_ft_SG}
    \end{subfigure}
    \caption{Distribution of $\log(p)$ of coreferent gender by antecedent gender in the PL and SG setting}
    \label{fig:gpt_ft}
\end{figure*}

\subsection{Models Fine-tuned with Gender-inclusive Language}

Figure \ref{fig:gpt_ft} presents the results for \citeposs{bartl-leavy-2024-showgirls} fine-tuned GPT-2 models. The models were fine-tuned for 3 epochs with an English corpus in which gendered terms were rewritten with gender-neutral variants and gendered singular pronouns (\textit{he, she}) were replaced with singular \textit{they}. 
The effects of pronoun replacement are clearly visible in the SG setting (Figure \ref{fig:gpt_ft_SG}): singular \textit{they} has a much higher likelihood than other pronouns that even overrides gender information from the antecedent. This indicates that fine-tuning may serve as a method for enabling models to accept singular \textit{they}, given that our findings demonstrate their difficulties with it (§\ref{sec:discussion}). However, the extent of replacement should likely be less comprehensive than in the experiments conducted by \citeposs{bartl-leavy-2024-showgirls}.
Further, in the PL setting, the probabilities resemble previously observed distributions for Qwen2.5 (Figure \ref{fig:qwen}) and OLMo (Figure \ref{fig:olmo}) for feminine and masculine antecedents. For neutral antecedents, however, masculine coreferents exhibit the highest probability, contrary to the intended effect of fine-tuning. We would have expected the fine-tuning process to enhance the likelihood of a neutral coreference and balance out associations between masculine and feminine coreferents. While fine-tuning with gender-neutral language might have been effective in reducing stereotyping \citep{bartl-leavy-2024-showgirls}, our results demonstrate that more fine-grained evaluation methods are necessary to comprehensively assess the effects.

\begin{figure*}[ht]
    \centering
    \includegraphics[width=\linewidth]{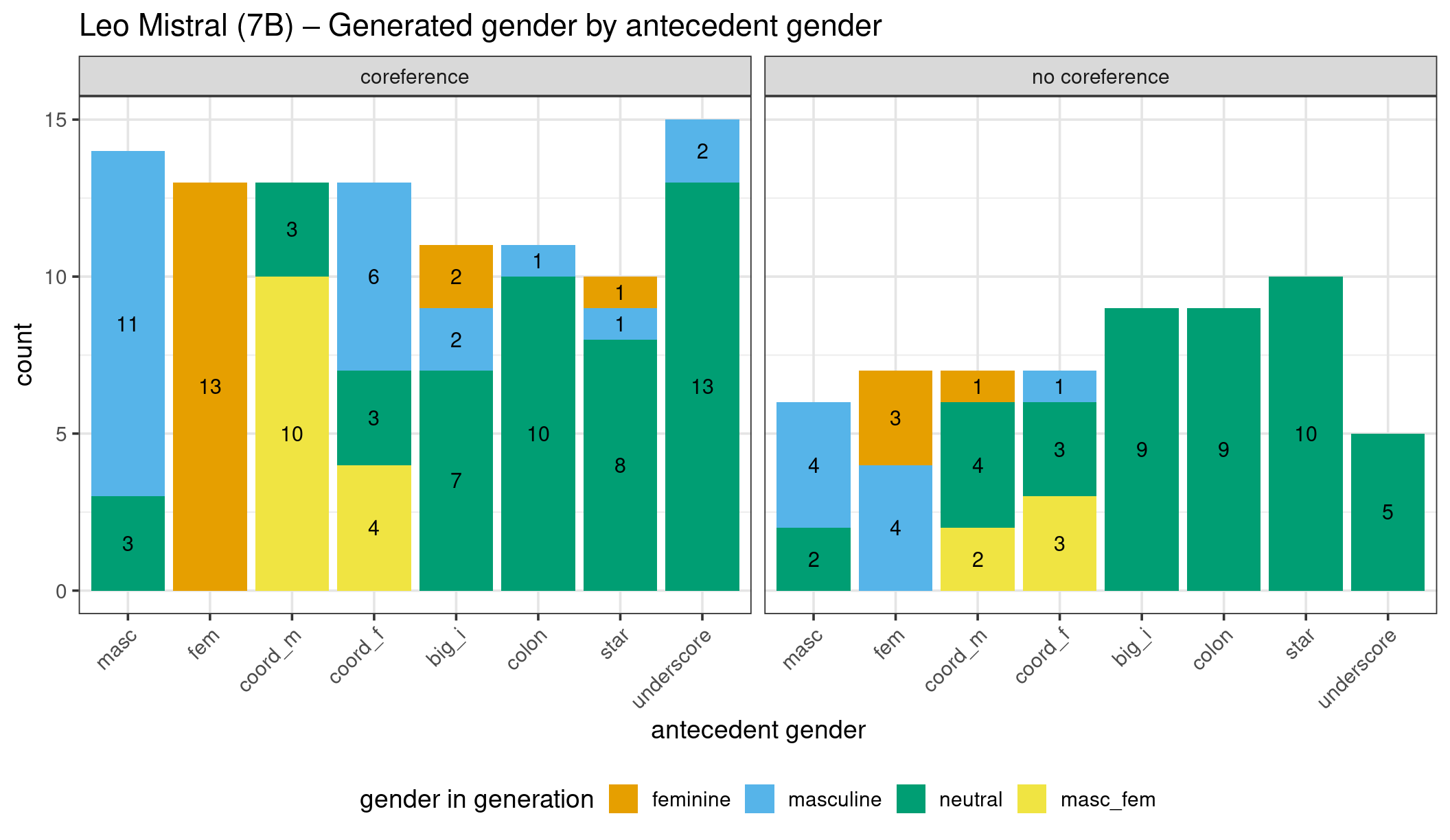}
    \caption{Generated gender for German model, divided by whether or not the continuation contains a coreferent of the antecedent}
    \label{fig:coref_gen_DE}
\end{figure*}

\subsection{German Coreferent Generation}

Figure \ref{fig:coref_gen_DE} visualizes how the different gender-inclusive strategies influence the gender mentioned in the generations. We differentiated by whether the model generation referred back to the antecedent (62.5\%, left panel) or not (37.5\%, right panel). What both conditions have in common is that in most cases gender-neutral antecedents effect a gender-neutral coreferent. For the \texttt{no coreference} group, indeed all coreferents are neutral. These results suggest that LLMs are likely to maintain gender-inclusive language when prompted with these forms. 
In fact, there were many instances in which the model simply repeated the antecedent phrase. This is why Figure \ref{ssec:coref_gen_DE} contains the additional coreferent category \texttt{masc\_fem} to capture instances in which the model generated coordinated forms (Table \ref{tab:gender_strategies_DE}, strategies 3\&4). These were indeed only generated when prompted with a coordinated form. 

For instances where the antecedent expressed only a single gender (masculine or feminine), Figure \ref{fig:coref_gen_DE} shows the majority of masculine coreferents and all of the feminine coreferents corresponding with the respective antecedent. These results indicate that feminine gender in the antecedent is a very strong signal for future generations. The presence of some neutral coreferents for masculine antecedents suggests that masculine gender can sometimes have a generic interpretation. However, in most cases masculine gender has a masculine association.

The German coreferent generation results suggest that generated coreferents generally align with the antecedent gender in the prompt, indicating that gender-inclusive language can encourage gender-neutral generations. However, this contrasts with our coreferent probability experiments (Section \ref{ssec:coref_prob_EN}), which revealed strong masculine biases. This suggests that German models rely on repetition rather than a genuinely gender-neutral interpretation.

\end{document}